\documentclass{article}

\usepackage[preprint]{neurips_2025}

\usepackage{amsmath}
\usepackage[utf8]{inputenc} 
\usepackage[T1]{fontenc}    
\usepackage{hyperref}       
\usepackage{url}            
\usepackage{booktabs}       
\usepackage{amsfonts}       
\usepackage{nicefrac}       
\usepackage{microtype}      
\usepackage{xcolor}         
\usepackage{multirow}
\usepackage{chngcntr}
\usepackage{graphicx}
\usepackage{multirow}
\usepackage{xcolor}
\usepackage{colortbl}
\usepackage{makecell} 
\usepackage{amssymb}
\usepackage{bbding}
\usepackage{pifont}
\usepackage{wasysym}
\usepackage{utfsym}
\usepackage{fontawesome}
\usepackage{wrapfig}
\usepackage{float}
\usepackage{authblk}

\definecolor{my_green}{RGB}{51,102,0}
\definecolor{my_red}{RGB}{204, 0, 0}
\renewcommand{\checkmark}{\textcolor{my_green}{\ding{51}}} 
\newcommand{\crossmark}{\textcolor{my_red}{\ding{55}}} 
\definecolor{ModelGreen}{RGB}{213,232,212}
\definecolor{ModelGrey_1}{RGB}{207,220,230}
\definecolor{ModelGrey_2}{RGB}{207,225,235}

\title{Task-Aware KV Compression For Cost-Effective Long Video Understanding }

\begin{document}

\author{
\textbf{Minghao Qin$^{1}$ \ \ Yan Shu$^{1,3}$ \ \ Peitian Zhang$^{1,4}$ \ \ Kun Lun$^{1,7}$ \ \ Huaying Yuan$^{1,4}$ \ \ Juenjie Zhou$^{1,5}$ \ \ Shitao Xiao$^{1}$ \ \ Bo Zhao$^{1,2}$ \ \ Zheng Liu$^{1,6†}$}

$^{1}$Beijing Academy of Artificial Intelligence  \; 
$^{2}$Shanghai Jiao Tong University \;
$^{3}$University of Trento \;
$^{4}$Renmin University of China \;
$^{5}$Beijing University of Posts and Telecommunications \;
$^{6}$Hong Kong Polytechnic University\;
$^{7}$Institute of Automation, CAS, Beijing, China}

\maketitle

\begin{abstract}
  Long-video understanding (LVU) remains a severe challenge for existing multimodal large language models (MLLMs), primarily due to the prohibitive computational cost. Recent approaches have explored KV compression to mitigate this issue, but they often suffer from significant information loss at high compression ratios. In this paper, we introduce \textbf{Video-X$^2$L}, which flexibly preserves critical video information for each LVU task. Video-X$^2$L involves two key operations. The first one is called \textit{bi-level KV compression}. During the MLLM's pre-filling stage, Video-X$^2$L generates two types of compressed KVs: low-compression KVs (L-KVs) to capture fine-grained video details and high-compression KVs (H-KVs) to offer compact video representations. The second one is called \textit{selective KV re-loading}. During the MLLM's decoding stage, Video-X$^2$L selectively re-loads L-KVs for the most critical video chunks while using H-KVs for other less important ones. This allows the MLLM to fully utilize task-specific information while maintaining the overall compactness. Video-X$^2$L is simple yet effective: it is free from additional training and directly compatible with existing KV-compressible MLLMs. We evaluate Video-X$^2$L with a variety of popular LVU benchmarks, including VideoMME, MLVU, LongVideoBench, and VNBench. Our experiment result shows that Video-X$^2$L outperforms existing KV-compression methods by a huge advantage while substantially saving the computation cost. Our source code has been made publicly available at the this repository\footnote{https://github.com/UnableToUseGit/VideoX22L}. 

\end{abstract}


\begin{figure*}
  \centering
  \includegraphics[width=0.90\textwidth]{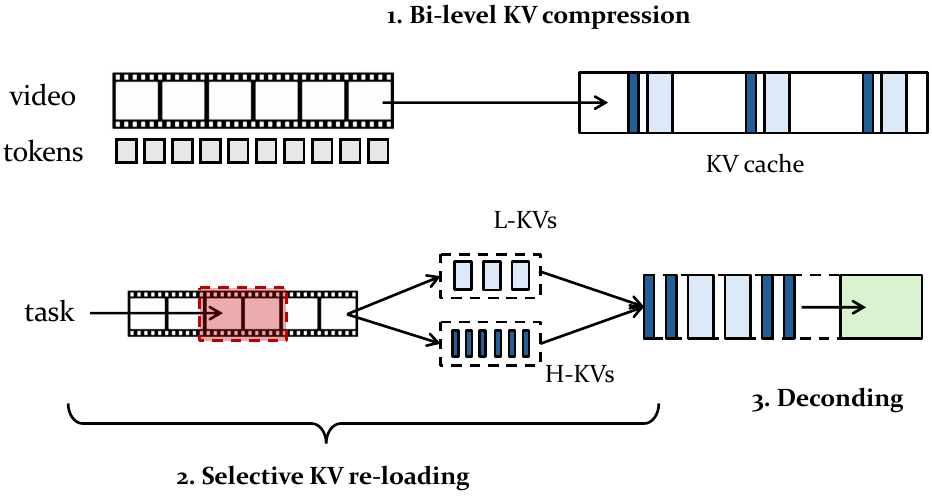}
  \vspace{1em} 
  \caption{The framework of Video-X$^2$L. First, it performs bi-level KV compression, which compresses KV cache into L-KVs (for fine-grained details) and H-KVs (for abstract representations). Second, it make selective use of the compressed KVs based on their importance to each task. The final answer is generated using the selectively re-loaded KVs of hybrid compression ratios.}  
  \label{fig:1}
\end{figure*}

\section{Introduction}



Multimodal Large Language Models (MLLMs) have made substantial progress in addressing tasks that require perception and reasoning over visual data~\cite{reid2024gemini,gpt4o,liu2023visual_llava}. Recently, there has been growing interest in extending MLLMs to video-related scenarios ~\cite{videollama,videollava,maaz2023videochatgpt,li2023videochat}, given their strong potential in modeling the spatial-temporal information within videos~\cite{videollama,videollava,maaz2023videochatgpt,li2023videochat}. However, existing MLLMs face significant challenges in handling long videos, primarily due to the prohibitive computational and memory costs. In particular, long videos consist of numerous frames, each producing a huge amount of visual tokens. As a result, the input often exceeds the model’s context window, making it computationally expensive and impractical for real-world applications. Even with extended context lengths, processing such a large number of visual tokens remains inefficient and resource-intensive. To alleviate these issues, various compression strategies have been explored to enable cost-effective processing of long videos. One common approach is token reduction, which removes redundant tokens before they are input into MLLMs~\cite{malmm2024,llamavid,maaz2023videochatgpt,moviechat2023}. Recently, KV compression has been introduced as a more promising strategy, which leverages MLLMs' inherent sparsification capability to generate compact representations for long videos~\cite{ye2024voco,shu2024video,liu2025video}. However, KV compression still suffers from severe information loss  especially under high compression ratios, limiting its ability to support tasks that require fine-grained analysis of long videos.

In this paper, we propose a novel framework for long-video understanding, called Video-X$^2$L. This method introduces two key operations to enhance the generation and utilization of compressed KVs (shown as Figure \ref{fig:1}). 

$\bullet$ \textbf{Bi-level KV compression}. Video-XL geneartes two types of compressed KVs during the pre-filling stage of MLLM: low-compression KVs (L-KVs) and high-compression KVs (H-KVs). Specifically, the input video is first partitioned into a series of chunks. While encoding each chunk, the original KVs from MLLMs are compressed as L-KVs using a small compression ratio, which preserves fine-grained information about the video. Meanwhile, the KVs are also compressed as H-KVs using a large compression ratio, resulting in highly abstract representations for the corresponding chunk. Both forms of KVs are cached in the main memory and reused by downstream tasks, thereby enabling the saving of computation cost.  

$\bullet$ \textbf{Selective KV re-loading}. Video-XL makes selective use of the compressed KVs during the decoding stage of MLLM. Given a specific LVU task, it first estimates the importance of each video chunk using a relevance oracle (e.g., a vision-language model, like CLIP, or simply the vanilla attention score from the MLLM). Based on this result, it recalls L-KVs for the most critical video chunks and H-KVs for other less important ones. The recalled KVs are re-ordered based on their relative position and re-loaded to the GPU memory. 

The decoding step is performed on top of the KVs with hybrid compression ratios. While the majority of video chunks are represented by H-KVs, the most critical ones are highlighted by L-KVs. This allows Video-X$^2$L to generate high-quality answers while maintaining strong cost-effectiveness. Additionally, Video-X$^2$L can be seamlessly integrated with existing KV-compressible MLLMs and arbitrary relevance oracles in a training-free way, making it extremely easy to be adopted in practice. 

We perform comprehensive experiments for Video-X$^2$L using a variety of popular LVU benchmarks, including VideoMME~\cite{videomme}, MLVU~\cite{zhou2024mlvu}, LongVideoBench~\cite{wu2024longvideobench}, and VNBench~\cite{vnbench}.  Our results show that Video-X$^2$L significantly enhances MLLM performance in long-video understanding across diverse scenarios. At the same time, it achieves substantial reductions in memory usage and notable improvements in inference speed, which enables more efficient processing of LVU tasks. Finally, the empirical advantage become more pronounced with higher compression ratios or longer video inputs, indicating Video-X$^2$L's potential in addressing the challenges posed by extra-long videos. 

To summarize, the contributions of this paper are highlighted as follows. 
\begin{itemize}
\item We propose a novel KV-compression framework Video-X$^2$L for LVU tasks for efficient and high-quality processing of LVU tasks. 
\item We introduce two key oeprations: bi-level KV compression and selective KV re-loading, to improve the generation and utilization of compressed KVs. 
\item We perform comprehensive experiment, verifying the effectiveness and efficiency of Video-X$^2$L in addressing the challenges associated with long-video understanding. 
\end{itemize}

\section{Related Work}
\label{sec:relatedwork}



\paragraph{Multimodal Large Language Models.}
Multimodal Large Language Models (MLLMs) are built upon Large Language Models (LLMs). They employ visual encoders to extract visual features and then utilize connectors to align these features with textual features, thereby enabling the model to effectively understand visual content (e.g., \cite{alayrac2022flamingo,li2023blip2,liu2023llava,gpt4o,reid2024gemini,zhu2023minigpt}). Over the past few years, MLLMs have achieved rapid development in both commercial and open-source domains. Commercial models, such as GPT-4o~\cite{gpt4o} and Gemini~\cite{reid2024gemini}, have long held a leading position in the field of multimodal understanding. However, recently, several excellent open-source MLLMs, including  DeepSeek-VL2~\cite{wu2024deepseek}, Qwen2.5-VL~\cite{bai2025qwen2} and InternVL3~\cite{zhu2025internvl3}, have emerged, demonstrating performance comparable to, or even surpassing, commercial models on many multimodal tasks. Despite these advances, most existing MLLMs~\cite{dai2024nvlm,tong2024cambrian,zhang2024llavanextvideo,liu2024llavanext,li2024llavaonevision,internvl,wang2024qwen2vl} primarily focus on understanding images or short video clips. Handling long multi-modal contexts remains a significant challenge, mainly due to the substantial resource overhead associated with memory and computation.

\paragraph{Video MLLMs.}
Building on the robust capabilities of Multimodal Large Language Models (MLLMs), which effectively integrate language understanding with visual processing, numerous studies~\cite{maaz2023videochatgpt, luo2023valley, stllm, li2023videochat, li2024mvbench, videollama, chatunivi} have since investigated the feasibility of extending their success from static image understanding to the dynamic domain of video analysis. However, the inherent temporal nature and high frame rate of video result in a substantial volume of visual tokens extracted by visual encoders. This sheer volume not only imposes significant computational and memory burdens for processing but also critically risks exceeding the limited context window of existing LLMs, a challenge that becomes particularly pronounced in Long Video Understanding (LVU). To address this challenge, most existing approaches employ external modules to reduce the number of visual tokens before they are fed into the LLM. For instance, MovieChat~\cite{moviechat2023} and MA-LMM~\cite{malmm2024} employ memory banks to filter and store key information, while LLaMA-VID~\cite{li2024llamavid} and Video-CCAM~\cite{fei2024videoccam} compress visual features into fixed query embeddings via cross-attention. Similarly, Oryx downsamples visual tokens through mean pooling and enhances feature interaction between pre-pooling and post-pooling stages using cross-attention. However, these methods, reduce visual token quantity before feeding them into LLM, inevitably result in serious visual information loss.  

In contrast, leveraging the KV sparsification characteristic of LLMs to compress visual information at the KV states level presents a more promising approach for handling long videos. VoCo-LLaMA~\cite{ye2024voco} pioneers this direction by compressing visual tokens into special tokens' KV states directly within LLM, demonstrating its effectiveness on benchmarks for both image and short video understanding. Video-XL~\cite{shu2024video} conducts an in-depth study on applying KV cache compression to LVU. By encoding video intervals sequentially, it compresses regular visual token into compact KV representations. This approach enables it to process up to 2048 frames on a single A100 GPU and achieve state-of-the-art (SOTA) performance on multiple mainstream LVU benchmarks. Although KV compression methods for LVU, such as Video-XL, achieves a remarkable performance, they still suffer from significant information loss under high compression ratios. To address this limitation, Video-X$^2$L proposes a simple yet effective solution: a non-uniform KV compression strategy combined with a task-aware reloading mechanism. This approach significantly enhances both performance and efficiency in LVU tasks.



\section{Video-X$^2$L}
\subsection{Preliminary}
Video-X$^2$L operates on top of a KV-compressible MLLM. Given an input video, the model first applies a visual encoder to tokenize the video into a long sequence of visual tokens: $X = [x_1, \dots, x_n]$. The visual tokens are subsequently encoded by the MLLM, which produces the KV cache $\{K, V\}$, where $|K|,|V| = n$, for downstream video understanding tasks. Since the vanilla KV cache can be prohibitively large for decoding, the MLLM further compresses $\{K, V\}$ into a more compact form: $\{K', V'\}$ for decoding, where $|K'|,|V'| = n' \ll n$. The compression ratio is defined as $\alpha = n/n'$. Prior approaches typically apply a uniform compression ratio across the entire video, resulting in an uniform compression of orginal KVs regardless of their underlying information. In theory, the compression yields an $\alpha\times$ reduction in memory usage and an $\alpha^2\times$ speedup in decoding. However, to fit long videos within the MLLM's limited context window, high compression ratios are often needed, which leads to a significant loss of critical information.  

\subsection{Overview}
Despite the association with long videos, the LVU tasks, such as causal reasoning and temporal grounding, exhibit non-uniform reliance on different parts of the video content. In most situations, it only needs to highlight a few crucial details while maintaining an abstract understanding of the global context (due to either redundancy or irrelevance). This property allows us to design flexible compression framework to alleviate the information loss from compression. In our framework (shown as Figure \ref{fig:1}), we perform bi-level KV compression while pre-filling the long-video input, with low-compression result (L-KVs) capturing fine-grained video details and high-compression result (H-KVs) offering abstract representation. Before decoding for a given task, it re-loads L-KVs for a few crucial parts of the video and H-KVs for the rest of less important parts. By doing so, useful information can be effectively preserved while maintaining the overall compactness. 

\subsection{Bi-level KV Compression}

Designed as a generic framework, Video-X$^2$L is applicable to any KV-compression MLLMs. In this place, we adopt the compression workflow from Video-XL~\cite{shu2024video} to illustrate its mechanism. Particularly, the KV compression is conducted w.r.t. each chunk of the long video. Given pre-defined chunk widths: $\{w_1,..., w_m\}$, the video tokens are partitioned into the following chunk list: 
\begin{equation}
    [ x_1, \dots , x_n ] \xrightarrow{\mathrm{Partition}} [X_1, \dots, X_m],
\end{equation}
where $|X_i| = w_i$ and $\sum_{1...m} w_i = n$. A group of visual summary tokens ($\mathcal{V}_i = \{vst_1,\dots,vst_{m_i}\}$) are uniformly inserted into each chunk ($X_i$), which reformulates the token sequence as follows: 
\begin{equation}
    X'_i \xleftarrow{\mathrm{ }} [X_{i,1}, \dots, X_{i,k}, vst_1, X_{i,k+1}, \dots, X_{i,2k}, vst_2, \dots, X_{i,mk}, vst_{m_i}].
\end{equation}
The visual summary tokens are introduced to compress the video information within each chunk. With the reformulation of each chunk, the compression-based self-attention is conducted:
\begin{equation}
    \mathrm{Atten}\big(\mathbf{Q}(X_i'),[\mathbf{K}(\mathcal{V}_{:i}), \mathbf{K}(X'_i)], [\mathbf{V}(\mathcal{V}_{:i}),\mathbf{V}(X'_i)]\big),
\end{equation} 
where $\mathbf{Q}(X'_i)$, $\mathbf{K}(X'_i)$, $\mathbf{V}(\mathcal{V}_{:i})$ are the QKVs of $X'_i$, $\mathbf{K}(X'_i)$, $\mathbf{V}(\mathcal{V}_{:i})$ are the KVs from previous visual summary tokens before $X'_i$, ``$[\cdot]$'' indicates the concatenation operator. The generated KVs of $\mathcal{V}_i$, i.e., $\mathbf{K}(\mathcal{V}_{i})$, $\mathbf{V}(\mathcal{V}_{i})$, are used as the KV compression result of $X'_i$. 

The compression ratio for each video chunk is controlled by the number of visual summary tokens, denoted as $m_i = |\mathcal{V}_i|$. Therefore, we introduce two different ratios for bi-level KV compression: a low compression ratio $\alpha_l$ and a high compression ratio $\alpha_h$, where $\alpha_l \ll \alpha_h$. The KV compression process is conducted for two passes. In the first pass, the low-compression KVs (L-KVs) are generated, with visual summary tokens densely placed in each video chunk to preserve fine-grained video details ($m_i = w_i/\alpha_l$). In the second pass, the high-compression KVs (H-KVs) are generated, where visual summary tokens are sparsely inserted into each video chunk ($m_i = w_i/\alpha_h$), yielding high abstract representations of the corresponding video chunk. Both L-KVs and H-KVs are off-loaded to CPU memory after the pre-filling stage, and are reused during decoding for the saving of computation cost. 

\subsection{Selective KV Re-loading} 

The compressed KVs are reloaded during decoding based on the estimated importance of each video chunk to a given LVU task. With loss of generality, the importance between the task $t$ and a video chunk $X_i$ is estimated by a relevance oracle: $s_{t,i} \leftarrow \mathrm{Rel}(t, X_i)$.  In practice, this oracle can take various forms, for instance, a specialized VLM, like LanguageBind~\cite{zhu2023languagebind}, or a simply using the aggregated attention scores from MLLMs~\cite{zhang2023h2o,xiao2024infllm}. Despite that different forms of importance estimators may generate distinct relevance result, our experiment shows that the proposed method well maintains a robust performance as long as a reasonably accurate estimation is made, probably due to the redundancy of information within KVs. 

With the estimation of importance, L-KVs are re-loaded to GPU memory for the most critical chunks, whose visual summary tokens are denoted as $\mathcal{V_A}$. Meanwhile, H-KVs are re-loaded for other less important chunks, with corresponding visual summary tokens denoted as $\mathcal{V_B}$. The re-loaded KVs are merged and re-ranged by the temporal order of visual summary tokens, leading to $\mathcal{V_{A+B}}$. As a result, the decoding operation is conducted based on the following context: 
\begin{equation}
    \hat{x}_{i} \leftarrow \mathrm{Decoding}(X_{<i}, X_{task}, \mathcal{V_{A+B}})
\end{equation}
where $\hat{x}_{i}$ is the $i$-th token to be predicted, $X_{<i}$ are the preceding tokens before $\hat{x}_{i}$, and $X_{task}$ are the tokens from the task's prompt. The tokens are assigned with their relative positions in the context while performing self-attention.

\section{Experiment}
\label{sec:results}
\subsection{Implementation}
We will presents a comprehensive evaluation of VideoX$^2$L, demonstrating its effectiveness and efficiency. Furthermore, we conduct ablation studies to investigate the impact of the selective mechanism and to explore whether appropriate post-training leads to performance improvements. To accomplish these objectives, our evaluation encompasses several widely recognized LVU benchmarks: MLVU~\cite{zhou2024mlvu}, VideoMME~\cite{videomme}, LongVideoBench~\cite{wu2024longvideobench}, and VNBench~\cite{vnbench}. The input lengths for these benchmarks are standardized as follows: 256 frames for MLVU, 128 frames for VideoMME and LongVideoBench, and 1 frame per second for VNBench. In addition to these benchmarks, we also utilize the Vision Needle-in-a-Haystack (V-NIAH) evaluation to assess VideoX$^2$L's performance under conditions of high compression and extended video lengths. Efficiency is quantified using Cache Size Reduction (\%) and TTFT (Time To First Token) as metrics, both recorded during the decoding phase. For the experiments detailed below, L-KV and H-KV are generated using compression ratios of 2× and 32×, respectively, unless stated otherwise. Each video undergoes initial uniform frame sampling into a sequence, subsequently partitioned into contiguous 10-frame chunks for subsequent processing. VideoX$^2$L's performance is optimized by tuning the top-k parameter of L-KVs selection for each task across all benchmarks, with k values ranging from 1 to 5. The VideoX$^2$L is designed to be highly adaptable, supporting various LVU models within the KV compression paradigm and different relevance oracles. To validate the effectiveness of our approach, we specifically employ Video-XL-7B~\cite{shu2024video} as the backbone and LanguageBind~\cite{zhu2023languagebind} as the relevance oracle in our experiments. All experiments are conducted on a machine with 8 A800-80GB GPUs.

\subsection{Benchmarks}
To comprehensively evaluate the effectiveness of our approach, our experiments utilize a diverse and comprehensive set of LVU benchmarks. These benchmarks offer varied perspectives on video characteristics and task types. Regarding video characteristics, MLVU, VideoMME, and LongVideoBench include videos spanning a significant range of durations, from minutes to hours. They also feature diverse content categories, such as sports, cartoons, movies, documents, and vlogs. In terms of task types, these three benchmarks respectively encompass 7, 12, and 17 distinct categories designed with different objectives. Unlike these, VNBench focuses specifically on assessing detail perception through needle tasks (retrieval, counting, and ordering). Collectively, the diversity across these benchmarks in video properties and task formulations provides a robust basis for validating our method's effectiveness. In addition to these benchmarks, we employ a Vision Needle-in-a-Haystack (V-NIAH) evaluation, leveraging QA pairs from VideoRope~\cite{wei2025videorope}, to further evaluate our framework's performance in extra-long scenes.

\begin{table*}[t]
\tiny
\centering
\vspace{-4mm}
\addtolength\tabcolsep{-2.4pt} 
\resizebox{1\linewidth}{!}{
\begin{tabular}{lcc|c|c|cc|c|c}
\toprule
\multicolumn{1}{c}{\multirow{2}{*}{Model}} & \multicolumn{1}{c}{\multirow{2}{*}{Size}} & \multicolumn{1}{c}{\multirow{2}{*}{Frames$^{\ast}$}} & \multicolumn{1}{c|}{MLVU Dev}  & \multicolumn{1}{c|}{MLVU Test}& \multicolumn{2}{c|}{VideoMME}& \multicolumn{1}{c|}{\multirow{2}{*}{LongVideo.}}  & \multicolumn{1}{c}{\multirow{2}{*}{VNBench}} \\

\multicolumn{1}{c}{} &  \multicolumn{1}{c}{} & \multicolumn{1}{c}{} & M-avg & M-avg & W/o sub & W sub & \multicolumn{1}{c|}{} & \multicolumn{1}{c}{} \\ 
\midrule
\rowcolor{gray!15}\multicolumn{9}{c}{\textbf{Proprietary Models}} \\
GPT-4V~\cite{openai2023gpt4}   & - & 256  & 49.2 & 43.3  & 59.5 & {63.3} & 59.1 & 48.9   \\
GPT-4o~\cite{gpt4o}  & - & 325 & \textbf{64.6} & \textbf{54.9} & 71.9  & {71.2} & \textbf{66.7} & 64.4 \\
Gemini-1.5-Pro~\cite{reid2024gemini} & - & 1010  & - &  -  &\textbf{75.0} &  {81.3} & 64.0 &  \textbf{66.7} \\
\midrule
\rowcolor{gray!15}\multicolumn{9}{c}{\textbf{Open-source MLLMs}} \\ 
VideoChat2~\cite{li2024mvbench} & 7B & 16  & 47.9 &35.1 & 39.5 & {43.8}  & 39.3  & 12.4  \\
LLaMA-VID~\cite{llamavid} & 7B & 651 & 33.2 &17.2 &- & {-} & - & 10.8 \\
VideoLLaVA~\cite{videollava} & 7B & 8 & 47.3 &30.7 & 39.9 & {41.6} & 39.1  & 12.4 \\
ST-LLM~\cite{stllm} & 7B & 64 & -  &37.9 & {42.3} & -  & 22.7 & - \\
Shargpt4Video~\cite{chen2024sharegpt4video} & 7B & 16 & 46.4 &33.8 & 39.9& {43.6}  & 39.7 & -\\
LLaVA-Next-Video~\cite{zhang2024llavanextvideo} & 34B & 16 & - &-  & 52.0& {54.9} & {50.5} & 20.1   \\
LongVA\dag~\cite{longva} & 7B & 256 & 56.3 & 41.1 &{52.6} & {54.3} &47.8& 47.8  \\
VideoLLaMA2\dag~\cite{cheng2024videollama2}& 8x7B & 32 &  - & -  &47.9 & {49.7} & 36.0 & 24.9  \\
Video-CCAM\dag~\cite{fei2024videoccam}& 9B & 96 & {58.5} & 42.9  &50.3 & {52.4} & 43.1 & 35.6\\
Long-LLaVA~\cite{wang2024longllava}& 13B & 64 & - &- & 51.9 & {-} & - & {52.1}\\

\(\text{Video-XL: }\text{2×}\) & 7B & 256 & \underline{64.6} & 42.6 & \underline{54.9} & \textbf{60.9} &  50.6 & \underline{63.0}  \\ 
\(\text{Video-XL: }\text{32×}\)  & 7B & 256 & 62.7 & \underline{44.4} & 52.9 & {59.2} & 49.0  & 52.9  \\ 
\midrule
\rowcolor{ModelGrey_2}\textbf{Video-X$^2$L} & 7B & 256 & \textbf{66.3} & \textbf{45.7} & \textbf{55.7} & \textbf{60.9} & \textbf{53.0} & \textbf{63.5}  \\ 
\bottomrule
\end{tabular}}
\caption{Experimental results on mainstream long video understanding benchmarks. Frames$^\ast$ indicates the maximum number of frames used by each model across all benchmarks.``LongVideo." refers to LongVideoBench. $\dag$ indicates that the results on VNBench and LongVideoBench were reproduced using their official weights.} 
\vspace{-4mm}
\label{tab:main_1} 
\end{table*}

\subsection{Main Results}


In this section, we present the performance of Video-X$^2$L on various LVU benchmarks alongside comparisons with both proprietary and open-source models.

As presented in Table~\ref{tab:main_1}, Video-X$^2$L demonstrates superior performance compared to all open-source MLLMs baselines considered. These baselines include models specialized for LVU, utilizing methods like token-level reduction, specific model architectures or training strategies, and KV compression. Among these, ``Video-XL: 2x" and ``Video-XL: 32x" respectively denote Video-XL that exclusively employ 2x low compression and 32x high compression\footnote{For a fair comparison, we did not use the performance results for Video-XL reported in its original paper due to the randomness in its original evaluation setting. The comparison results and explanation can be found in the Appendix A.1.}

Focusing on VNBench, which utilizes synthetic video samples and emphasizes needle tasks such as counting and ordering, our method yields leading performance. In contrast to VNBench, the remaining three benchmarks—MLVU, VideoMME, and LongVideoBench—feature significantly longer video durations, spanning minutes to hours, and encompass a broader array of task categories. Video-X$^2$L still maintains better or comparable performance across these benchmarks compared to baselines. It is noteworthy that, compared to ``Video-XL: 2x", the performance gain is substantial on MLVU and LongVideoBench (+1.7 and +2.5), whereas the improvement on VideoMME is comparatively less pronounced (+0.8). This disparity can be attributed to the nature of the tasks: MLVU and LongVideoBench primarily consist of detailed QA tasks with content referencing, while many QA tasks within VideoMME incorporate concrete or ambiguous timestamps, which our relevance oracle struggles to handle effectively. Additionally, from an efficiency perspective, Video-X$^2$L uses only a small portion of low-compression KVs (L-KVs) but achieves significant performance enhancement compared to ``Video-XL: 2x" and ``Video-XL: 32x", further demonstrating our method's effectiveness and efficiency.

Overall, Video-X$^2$L leverages Bi-level KV compression and a selective KVs reloading mechanism to achieve superior performance, outperforming both Video-XL and other models on various scenarios.

\begin{table*}[t]
\tiny
\centering
\addtolength\tabcolsep{-2.4pt} 
\resizebox{0.9\linewidth}{!}{
\begin{tabular}{l|cccc|c|c|c}
\toprule
\multicolumn{1}{c|}{\multirow{2}{*}{Model}}  & \multicolumn{4}{c|}{MLVU} & \multicolumn{1}{c|}{Cache Size} & \multicolumn{1}{c|}{\multirow{2}{*}{TTFT (ms)}} & \multicolumn{1}{c}{\multirow{2}{*}{Speedup}} \\ 
{} & Holistic & Single. & Multi. & M-avg & {Reduction (\%)} & {} & {} \\ 
\midrule
\rowcolor{gray!15}\multicolumn{8}{c}{\textbf{Baselines}} \\
\text{Video-XL}: \text{w/o cmpr} & 69.4 & 63.7 & 50.1 & 62.0 & 00.0\% & 233.1 & 1.0× \\
\text{Video-XL}: \text{2×} & 70.5 & 66.3 & 54.4 & 64.6 & 50.0\% & 139.7 & 1.7× \\
\text{Video-XL}: \text{4×} & 68.5 & 67.3 & 54.2 & 64.8 & 75.0\% & 73.7 & 3.2× \\
\midrule
\rowcolor{gray!15}\multicolumn{8}{c}{\textbf{Group 1: 2× low with (8×, 16×, 32×, 72×) high compression }} \\
\(\text{Video-X$^2$L: }\text{2×8×}\) & 68.3 & 67.2 & 55.7 & 65.0 & 78.8\% & 65.8  & 3.5× \\
\(\text{Video-X$^2$L: }\text{2×16×}\) & 69.0 & 67.5 & 56.3 & 65.4 & 83.7\% & 56.3 & 4.1× \\
\rowcolor{ModelGrey_2}\(\textbf{Video-X$^2$L: }\textbf{2×32×}\) & 70.3 & \textbf{67.8} & \textbf{56.8} & \textbf{66.0} & 86.1\% & 52.1 & 4.5× \\
\(\text{Video-X$^2$L: }\text{2×72×}\) & \textbf{70.7} & 67.4 & 54.4 & 65.3 & \textbf{87.4\%} & \textbf{51.4} & \textbf{4.5×} \\
\midrule
\rowcolor{gray!15}\multicolumn{8}{c}{\textbf{Group 2: 4× low with (8×, 16×, 32×, 72×) high compression }} \\
\(\text{Video-X$^2$L: }\text{4×8×}\) & 69.4 & \textbf{67.7} & 54.2 & 65.2 & 84.6\% & 55.2 & 4.2× \\
\(\text{Video-X$^2$L: }\text{4×16×}\) & 69.2 & 67.3 & 55.1 & 65.0 & 89.4\% & 48.1 & 4.8× \\
\rowcolor{ModelGrey_2}\(\textbf{Video-X$^2$L: }\textbf{4×32×}\) & \textbf{70.0} & 67.5 & \textbf{55.3} & \textbf{65.4} & 91.8\% & 42.9 & 5.4× \\
\(\text{Video-X$^2$L: }\text{4×72×}\) & 69.4 & 67.1 & 52.5 & 64.5 & \textbf{93.2\%} & \textbf{40.2} & \textbf{5.8×} \\
\bottomrule
\end{tabular}}
\caption{Impact of different compressions ratios combination on task performance and efficiency. ``\(\text{Video-XL: }\text{m×n×}\)" uses a low compression ratio of m and a high compression ratio of n. ``\(\text{Video-XL: }\text{w/o cmpr}\)" does not use the compression mechanism and inference based on regular visual tokens. ``Single." refers to single-detail tasks. ``Multi." refers to multi-detail tasks. ``TTFT" refers to the time of first token generated in decode phase.} 
\label{tab:main_2} 
\end{table*}

\subsection{Cost-effective Analysis} 

In the previous section, we primarily discussed the strong performance of Video-X$^2$L across various benchmark. In this section, we will further explore in detail the significant efficiency enhancement it brings and its effectiveness on various concrete task types.

We conduct experiments on holistic, single-detail and multi-detail tasks from MLVU. We establish three baselines: (1) Video-XL without compression (Video-XL: w/o cmpr), (2) Video-XL with a low compression ratio of 2× (Video-XL: 2×), and (3) Video-XL with a low compression ratio of 4× (Video-XL: 4×). To comprehensively demonstrate and discuss Video-X$^2$L's effectiveness and efficiency, we evaluate Video-X$^2$L under diverse compression ratio combination. Specifically, we paired high compression ratios (2×, 4×) with low compression ratios (8×, 16×, 32×, 72×) to construct various bi-level KVs, resulting in two distinct combination groups for experimental validation. All settings reload top-3 task-relevant L-KVs and employ the default naive attention mechanism to ensure a fair comparison.


As shown in Table~\ref{tab:main_2}, it is first notable that Video-X$^2$L with various compression configurations all achieve outstanding efficiency gains compared to ``Video-XL: w/o cmpr". The most significant improvement is observed with the ``Video-X$^2$L: 4×72×" setting, achieving a \textbf{93.2\%} KV cache reduction and a \textbf{5.8×} decoding speedup. Furthermore, compared to Video-XL with fixed low-compression ratios (2×, 4×), Video-X$^2$L still demonstrates clear efficiency advantages. Beyond its outstanding efficiency, Video-X$^2$L also demonstrates superior performance, surpassing all baselines across various compression ratio combinations, except for ``Video-XL: 4×72×", whose performance is slightly lower than the baseline (with a marginal drop of 0.3 points). The optimal setting is ``Video-X$^2$L 2× 32×", which achieves a 1.4-point improvement compared to the baseline applied the same low compression ratio.

Further analysis, we observe that our approach is particularly well-suited for detail-oriented tasks. The performance on both single-detail and multi-detail tasks in both groups is either better than or comparable to the corresponding baselines, with the exception of ``Video-XL: 4×72×". For holistic tasks, VideoX$^2$L doesn't show an obvious advantage. However, the flexible selective KV reloading mechanism allows VideoX$^2$L to consistently load L-KVs for the initial and final chunks of the video. This approach provides sufficient global information to ensure that VideoX$^2$L maintains comparable performance on holistic tasks.

Additionally, in both Group 1 and Group 2, we observe a general trend where the mean average score (M-avg) initially increases and then decreases as the high compression ratio increases. This initial increase can be explained by the reduction of noise in the H-KVs as the compression ratio grows. With fewer distractions, the model can better understand global context and focus on fine-grained information preserved in the L-KVs. However, when the high compression ratio reaches 72×, both groups experience a significant performance drop compared to 32×. This phenomenon can be attributed to two main factors. First, due to the inherent limitations of the relevance oracle's capability, some relevant and valuable information is incorrectly reloaded with H-KVs. As the high compression ratio increases, the loss of critical information becomes more pronounced. Second, excessive compression of less relevant video chunks disrupts the complete video context, leading to the loss of valuable global information that is essential for maintaining overall coherence and understanding. Therefore, selecting an appropriate high compression ratio is crucial to preserve global information and compensate for the limitations of the relevance oracle's capability.

\begin{figure*}[t]
    \vspace{-4mm}
   \centering
   \includegraphics[width=1.0\linewidth]{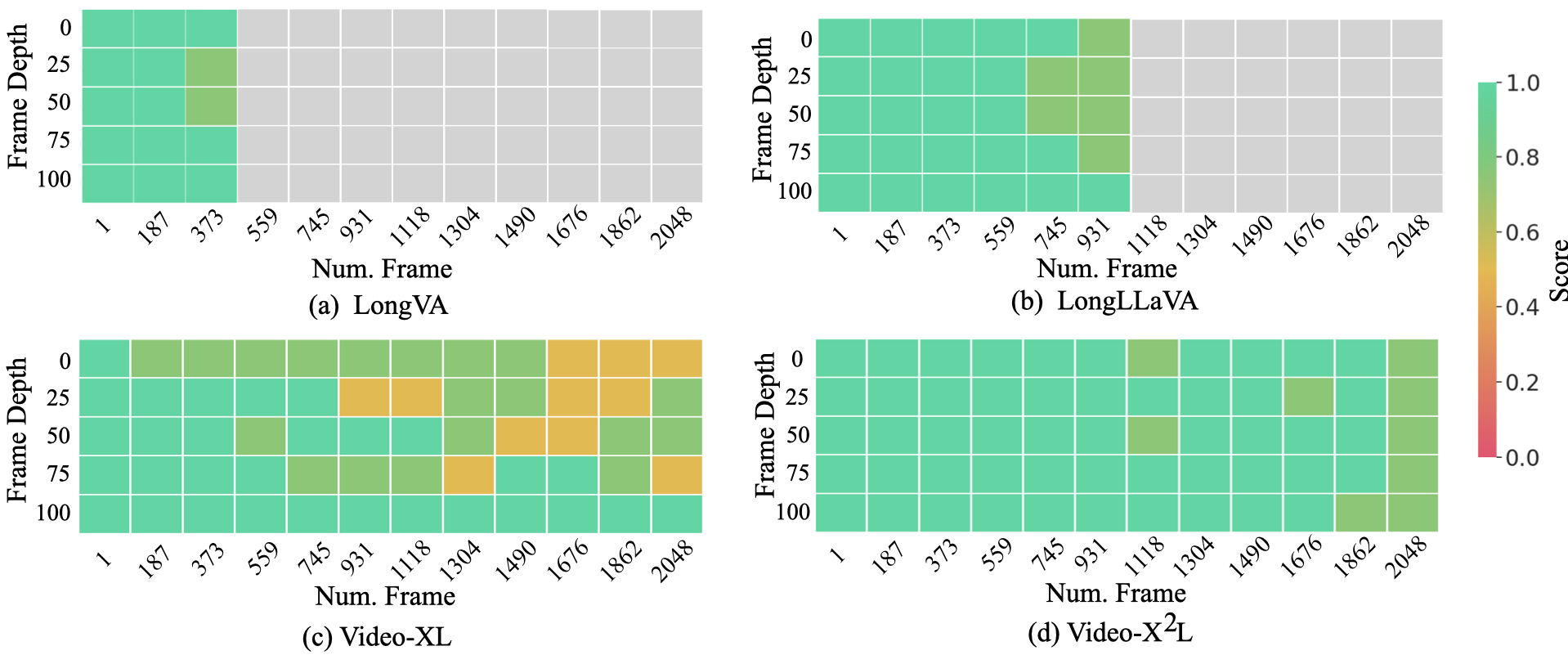}
   \caption{Results on the Needle-In-A-Haystack evaluatio within a single A100 80GB GPU. The x-axis represents the total number of frames in the video haystack. The y-axis shows the position where the needle image is located.} 
   \label{fig:exp3}
   \vspace{-4mm}
\end{figure*}

\subsection{Extra-Long LVU Evaluation with NIAH}
When processing extra-long videos, compression-based LVU models must employ high compression ratios due to resource constraints. This often results in substantial information loss, specifically impacting fine-grained details, predominantly within the central portions of the video. Nevertheless, VideoX$^2$L is particularly well-suited for such extra-long video settings. This is attributed to its low resource demanding and robust detail perception capabilities, conferred by its bi-level KVs and selective KV reloading mechanism. We demonstrate this by conducting a Needle-In-A-Haystack (NIAH) evaluation~\cite{longva} on an A100-80GB GPU. We compare Video-X$^2$L against three representative LVU MLLMs: LongVA, LongLLaVA, and Video-XL. Each of these models employs a distinct approach: LongVA transfers LLMs' long-context understanding to video understanding through specialized training, LongLLaVA modifies the model architecture to handle extended contexts, and Video-XL utilizes KV compression technique for efficient long video processing. As illustrated in Figure~\ref{fig:exp3}, Video-X$^2$L demonstrates superior performance in both input length and response accuracy compared to the baselines. Specifically, it supports up to 2048 input frames, outperforming LongVA and LongLLaVA, while maintaining 100\% accuracy for inputs below 1000 frames. Compared to Video-XL, Video-X$^2$L achieves significantly better performance beyond 1000 frames. This is because the two key innovative mechanism make Video-X$^2$L avoid suffering information loss under high compression in the extra-long scenario. This is because the two key innovative mechanisms prevent VideoX$^2$L from suffering severe information loss under high compression in extra-long scenarios.

\subsection{Ablation Studies}
We conducted extensive ablation studies to explore the impact of relevance oracle types, post-training, top-k selection and bi-level KVs. The last two experiments can be found in Appendix A.2 and A.3.


\begin{wraptable}{r}{0.5\textwidth}  
\centering
\small
\begin{tabular}{l|c|c}
\toprule
Settings & MLVU & LongVideo. \\ 
\midrule
Video-XL: 2× & 64.6 & 50.6 \\
Random & {63.7}  & {50.3} \\
Heuristic: Lats-N & {64.0}  & {48.9} \\
Heuristic: Uniform & {64.0}  & {50.6} \\
\midrule
\text{Attention Score} & \text{64.9}  & {48.9} \\
\text{SigLip} & \text{65.4}  & {51.2} \\
$\text{InternVideo2}_\text{1B}$ & {65.2}  & {51.5} \\
$\text{InternVideo2}_\text{6B}$ & {65.4}  & \text{51.8} \\
\rowcolor{ModelGrey_2}LanguageBind & \textbf{65.4} &  \textbf{51.7} \\
\bottomrule
\end{tabular}
\caption{Analysis of the effectiveness of various relevance oracle at top-3 relevant chunks selected.} 
\vspace{-3mm}
\label{tab:abs3} 
\end{wraptable}

\textbf{Relevance Oracle Types} We have previously demonstrated the strong performance and high efficiency of VideoX$^2$L. These impressive results are attributable to its mechanism design, which alleviates the need for additional training for both the model and the relevance oracles. This inherent property renders our approach highly adaptable, facilitating its seamless integration with various selective mechanism in practice. To empirically validate this capability, we conduct an evaluation employing distinct selective mechanisms, including attention score based selection~\cite{shen2024longvu}, and various multi-modal retrieval models (LanguageBind~\cite{zhu2023languagebind}, InternVideo2$_{1B}$, InternVideo2$_{6B}$~\cite{wang2024internvideo2}, and SigLip~\cite{tschannen2025siglip}). The first selective mechanism computes cross-attention between the task query's LLM embedding vector and each frame feature have aligned by MLP projector. This calculation yields frame-level relevance scores with respect to the task query. A chunk-level relevance score is then obtained by averaging these frame-level scores within the chunk. L-KVs are selected and loaded based on these chunk-level scores. For these retrieval models, LanguageBind and InternVideo2 are classic CLIP-style video retrievers. Meanwhile, SigLip is a different architecture vision-language model primarily focus on image multimode understanding. Our choices are diverse in backbone type, model size and task focus. Furthermore, we establish four baseline methods for comparison: (1)Video-XL: 2×, utilizing a lower compression ratio of 2. (2)Random: Randomly selects and loads L-KVs. (3)Last N: Selects and loads the last N video chunks of L-KVs. (4)Uniform: Selects and loads L-KVs at fixed equal intervals. As illustrated in Table~\ref{tab:abs3}, VideoX$^2$L achieves consistently superior performance when integrated with various relevance oracles, outperforming all baselines (Attention score based selection achieve better performance on MLVU.). This empirically validates that our approach is simple, effective, and readily applicable, compatible with diverse selective mechanisms. Moreover, results suggest that dedicated multi-modal retrievers offer improved performance over simpler attention score-based selection methods, positioning them as a more advantageous choice.

\begin{wraptable}{r}{0.6\textwidth}
\small
\centering
\begin{tabular}{l|c|c|c}
\toprule
Model & Post-training & MLVU & VideoMME \\ 
\midrule
\(\text{Video-XL: }\text{2×}\) & \crossmark  & 64.6 & 54.9 \\
Video-X$^2$L & \crossmark  & 66.3 (\(\uparrow 1.7\)) & 55.7 (\(\uparrow 0.8\)) \\
\midrule
\(\text{Video-XL: }\text{2×}\) & \checkmark  & {64.6} & {54.8}  \\
Video-X$^2$L & \checkmark  & {66.7 (\(\uparrow 2.1\))} & {56.0 (\(\uparrow 1.1\))}  \\
\bottomrule
\end{tabular}
\caption{Analysis of the impact of post-training.} 
\vspace{-3mm}
\label{tab:abs_2} 
\end{wraptable}

\textbf{Post-training}. Our experiments have shown that Video-X$^2$L, a training-free method, significantly improves performance via bi-level KVs and selective KV reloading mechanism. In fact, our approach can get a further improvement only need a lightweight post-training to enhances its adaptation to these mechanisms. To validate this, we conduct an ablation study using 4K video QA data sampled from Video-XL's training set, ensuring minimal gains in visual understanding. Both Video-XL and Video-X$^2$L are trained under the same hyperparameters, with Video-X$^2$L reloading top-3 task-relevant KVs and using a (2×, 32×) compression ratio, while Video-XL uses only 2× compression. As shown in Table~\ref{tab:abs_2}, post-training significantly widens the performance gap between Video-XL and Video-X$^2$L on MLVU and Video-MME, confirming that Video-X$^2$L better adapts to the two mechanisms with minimal post-training.




\section{Conclusion}
\label{sec:conclusion}
In this paper, we proposed Video-X$^2$L, a novel framework for long-video understanding (LVU) that integrates bi-level key-value (KV) compression with a selective KV re-loading mechanism. By dynamically reloading low-compression KVs for relevant video segments and high-compression KVs for less critical ones, Video-X$^2$L effectively captures essential information across long videos. Extensive experiments on popular LVU benchmarks demonstrate that Video-X$^2$L significantly outperforms existing methods while substantially reducing the computational cost. 


{
\small
\bibliographystyle{plain}
\bibliography{main}
}

\newpage
\appendix
\section{Appendix} 

\subsection{Additional Comparison with Video-XL}
\begin{table}[H]
\tiny
\centering
\addtolength\tabcolsep{-2.4pt} 
\resizebox{1\linewidth}{!}{
\begin{tabular}{lcc|c|c|cc|c|c}
\toprule
\multicolumn{1}{c}{\multirow{2}{*}{Model}} & \multicolumn{1}{c}{\multirow{2}{*}{Size}} & \multicolumn{1}{c}{\multirow{2}{*}{Frames$^{\ast}$}} & \multicolumn{1}{c|}{MLVU Dev}  & \multicolumn{1}{c|}{MLVU Test}& \multicolumn{2}{c|}{VideoMME}& \multicolumn{1}{c|}{\multirow{2}{*}{LongVideo.}}  & \multicolumn{1}{c}{\multirow{2}{*}{VNBench}} \\
\multicolumn{1}{c}{} &  \multicolumn{1}{c}{} & \multicolumn{1}{c}{} & M-avg & M-avg & W/o sub & W sub & \multicolumn{1}{c|}{} & \multicolumn{1}{c}{} \\ 
\midrule
\(\text{Video-XL: }\text{2×}\) & 7B & 256 & {64.6} & 42.6 & 54.9 & \underline{60.9} &  50.6 & \underline{63.0}  \\ 
\(\text{Video-XL: }\text{32×}\)  & 7B & 256 & 62.7 & 44.4 & 52.9 & {59.2} & 49.0  & 52.9  \\ 
\(\text{Video-XL}\) & 7B & 256 & \underline{64.9} & \underline{45.5} & \underline{55.5} & \textbf{61.0} & \underline{50.7}  & 61.6  \\ 
\midrule
\rowcolor{ModelGrey_2}\textbf{Video-X$^2$L} & 7B & 256 & \textbf{66.3} & \textbf{45.7} & \textbf{55.7} & \underline{60.9} & \textbf{53.0} & \textbf{63.5}  \\ 
\bottomrule
\end{tabular}}
\caption{The comparison between Video-XL original evaluation and our approach} 
\vspace{-6mm}
\label{tab:appen1} 
\end{table}

The original evaluation of Video-XL was conducted using a mix of random compression ratios. Consequently, this approach resulted in unstable performance across different benchmarks. As illustrated in Table~\ref{tab:appen1}, this instability is evident, with Video-XL's performance occasionally surpassing ``Video-XL:2×", being comparable at other times, and sometimes being weaker. Therefore, to facilitate a clearer and fairer comparison, we standardized the compression ratio for Video-XL. Despite being compared against its original reported performance, VideoX$^2$L still demonstrates superior performance.

\subsection{Top-K Selection and Reloading}

\begin{figure}[H]
    \centering
    \includegraphics[width=0.7\linewidth]{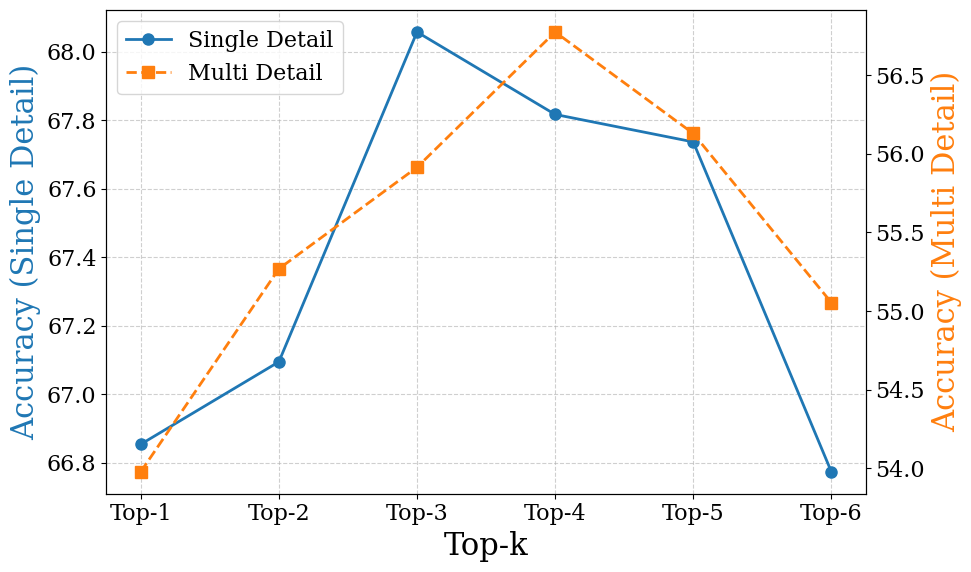}
    \caption{Analysis on the impact of recall@topk.}
    \vspace{-2mm}
    \label{fig:abs4}
\end{figure}

To investigate the impact of the top-\(k\) selection and reloading task-relevant L-KVs on Video-X$^2$L's performance, we conduct comparative experiments with different values of \(k\) ranging from 1 to 6. We evaluate the model on single-detail and multi-detail tasks from MLVU. As shown in Figure~\ref{fig:abs4}, the accuracy for both tasks initially improves and then declines as \(k\) increases. This trend suggests that, at lower values of \(k\), increasing the number of selected and reloaded L-KVs enhances the recall rate of the relevance oracle, providing the model with more relevant fine-grained information and improving its performance. However, as \(k\) continues to grow, the inclusion of increasingly irrelevant yet fine-grained information introduces noise, ultimately degrading the model's accuracy. This observation highlights the importance of balancing the quantity and relevance of reloaded KVs for optimal performance.

\subsection{Effectiveness of L-KVs and H-KVs} 

\begin{table}[H]
\tiny
\centering
\addtolength\tabcolsep{-2.4pt} 
\resizebox{0.7\linewidth}{!}{
\begin{tabular}{l|c|c|c}
\toprule
Model & MLVU & VideoMME & LongVideo. \\ 
\midrule
w/o high compression & 62.7 & 52.9 & 51.2  \\
w/o low compression & 62.4 & 52.4 & 49.0 \\
\rowcolor{ModelGrey_2} Video-X$^2$L & \textbf{65.4 } & \textbf{54.3} & \textbf{51.7} \\
\bottomrule
\end{tabular}}
\vspace{2em}
\caption{Analysis of the impact of L-KVs and H-KVs at top-3 relevant chunks selected.} 
\label{tab:abs_1} 
\end{table}

To further evaluate the effectiveness of the two types of KVs, we conduct a comparative study across MLVU, VideoMME, and LongVideoBench. In all evaluations, we select and reload the top-3 task-relevant H-KVs. For comparison, we establish two configurations: (1) no L-KVs are reloaded for task-relevant chunks (denoted as w/o low-compression''), and (2) only L-KVs are retained (denoted as w/o high-compression''). As demonstrated in Table~\ref{tab:abs_1}, the performance degradation observed across all three benchmarks in the absence of either type of KVs highlights the critical importance of bi-level KVs mechanism.

\section{Discussion On Limitations}
While Video-X$^2$L has made remarkable advancements in long video understanding, there remain several opportunities for further improvement.First, VideoX$^2$L performs well on detailed QA tasks, with the exception of QA tasks with timestamps, which are difficult for current relevance oracles to perceive and locate. Second, our approach is efficient during the decoding stage, but it lacks acceleration in the pre-filling stage. We plan to explore these directions in our future research.

\end{document}